\documentclass[conference]{IEEEtran}
\IEEEoverridecommandlockouts

\usepackage[utf8]{inputenc}
\usepackage{cite}
\usepackage{amsmath,amssymb}
\usepackage{graphicx}
\usepackage{textcomp}
\usepackage{xcolor}
\usepackage{booktabs}
\usepackage{float}

\usepackage{tikz}
\usetikzlibrary{shapes.geometric,arrows.meta,positioning}

\usepackage[caption=false,font=footnotesize]{subfig}

\usepackage{hyperref}

\tikzset{
  process/.style = {
    rectangle,
    draw,
    rounded corners,
    text centered,
    minimum height=1cm,
    text width=10cm,
    font=\small
  },
  arrow/.style = {
    thick,
    -{Stealth}
  }
}

\def\BibTeX{{\rm B\kern-.05em{\sc i\kern-.025em b}\kern-.08em
    T\kern-.1667em\lower.7ex\hbox{E}\kern-.125emX}}

\title{Detecting AI-Generated Images via Diffusion Snap-Back Reconstruction: A Forensic Approach}

\author{
\IEEEauthorblockN{
Mohd Ruhul Ameen\textsuperscript{1}, 
Akif Islam\textsuperscript{2}
}
\IEEEauthorblockA{
\textsuperscript{1,2}Department of Computer Science and Engineering, University of Rajshahi, Rajshahi, Bangladesh\\
\textsuperscript{1}ameensunny242@ru.ac.bd, 
\textsuperscript{2}iamakifislam@gmail.com
}
}

\begin{document}

\maketitle

\begin{center}
\small Accepted at the 2026 IEEE 2nd International Conference on Quantum Photonics, Artificial Intelligence and Networking (QPAIN 2026). Copyright IEEE.
\end{center}

\begin{abstract}
The rapid advancement of generative image models has transformed digital media to the point where AI-generated images can no longer be reliably distinguished from authentic photographs by human observers or many conventional detection methods. Modern text-to-image systems such as Stable Diffusion and DALL-E can now generate images so realistic that they often appear completely natural, leaving little to no visible artifacts for traditional deepfake detectors to rely on. This challenge has practical consequences for misinformation control, institutional identity verification, and digital trust in political and legal contexts. Instead of searching for hidden pixel-level traces, we take a different approach: we observe how an image responds when it is gently disturbed and reconstructed by a diffusion model. We call this behavior \textit{diffusion snap-back}. By tracking how perceptual similarity measures (LPIPS, SSIM, and PSNR) change across different reconstruction strengths, we capture compact and interpretable signals that reveal how closely an image aligns with the diffusion model’s learned denoising behavior. Evaluated on a balanced dataset of 4,000 human and AI-generated images, the proposed method achieves an AUROC of 0.993 under stratified five-fold cross-validation and 0.990 on a holdout split using only logistic regression. Initial robustness tests show that the method remains stable under common real-world distortions such as image compression and added noise. Although our experiments were conducted using a single diffusion backbone, the results indicate that reconstruction behavior can serve as a reliable and scalable foundation for synthetic media detection as generative models continue to grow more realistic.
\end{abstract}


\section{Introduction}
\label{sec:intro}

Large-scale generative models have transformed how visual content is created and consumed. Modern diffusion systems such as Stable Diffusion, Midjourney, DALL-E and Gemini 3.1 Flash Image can synthesize highly realistic images across diverse domains~\cite{cao2022survey,zhang2023survey}. Recent multimodal tools capable of producing near-photorealistic video further blur the boundary between authentic and synthetic media~\cite{chow2025veo3}. As visual artifacts disappear, distinguishing real from generated content becomes increasingly difficult.

This technological shift carries serious societal and institutional risks. AI-generated visuals are now widely used in misinformation campaigns, political propaganda, and reputational attacks~\cite{dismislab2025bangladesh}. Older or less digitally literate populations may be particularly vulnerable~\cite{akter2024awareness}. Beyond public misinformation, synthetic imagery also raises concerns in everyday institutional settings. University or academic admission systems that rely on uploaded photographs may be vulnerable to impersonation or proxy examination schemes, where a synthetic or manipulated image is submitted online and a different individual later appears for the exam. Fabricated visuals can also influence criminal investigations or legal disputes, and manipulated images may fuel rumors in highly polarized political environments. These challenges highlight the need for scalable and reliable authenticity verification tools.

Unlike traditional deepfakes, diffusion-based systems generate entirely new scenes from learned distributions~\cite{moser2024diffusionSurvey}. Consequently, many legacy detection methods—such as frequency-based artifact analysis or GAN-specific fingerprinting—fail to generalize. Diffusion-generated images often lack obvious statistical irregularities, motivating a shift from static artifact inspection to behavior-based analysis.

In this work, we examine how images respond when reconstructed by a diffusion model under controlled noise strengths $\mathcal{S}={0.15,0.30,0.60,0.90}$. Diffusion models refine noisy inputs through iterative denoising, and the way an image reconstruct reveals how closely it aligns with the model’s learned prior. We observe that AI-generated images typically degrade more smoothly as reconstruction strength increases, whereas authentic photographs often show sharper perceptual changes at higher noise levels. Rather than claiming strict separation, we interpret this difference as varying degrees of alignment with the denoising prior. We term this reconstruction behavior \emph{diffusion snap-back}, which forms the basis of our forensic framework. The contributions of this work are as follows:

\begin{enumerate}
  \item We introduce a reconstruction-dynamics detection framework that treats a pre-trained diffusion img2img pipeline as a forensic probe rather than relying on pixel-level artifacts.
  \item We design a compact 15-dimensional feature representation combining multi-strength LPIPS, SSIM, and PSNR metrics with trajectory descriptors (AUC-LPIPS, $\Delta_{LP}$, knee-step).
  \item We establish a lightweight classification pipeline based on logistic regression and stratified cross-validation to operationalize diffusion snap-back features for synthetic image detection.
  \item We conduct ablation, correlation, and robustness analyses to assess feature contribution, interpretability, and stability under common distortions.
\end{enumerate}

\section{Related Work}
\label{sec:related}

\subsection{Synthetic Image Detection}  
Early efforts in synthetic media detection primarily focused on GAN-generated images, particularly facial content. Initial approaches leveraged frequency-domain anomalies or trained CNN classifiers to detect GAN-specific artifacts~\cite{karras2019style, goodfellow2014generative}. Although effective in constrained settings, these methods often fail to generalize beyond face-centric datasets. Recent surveys report that many GAN-era detectors struggle when applied to newer generative models or broader visual domains~\cite{dogoulis2023improving, ojha2023towards}. The rise of diffusion-based generators has further reduced visible frequency artifacts and introduced smoother textures with stronger physical consistency, weakening assumptions underlying legacy detectors~\cite{chu2025fire}. Additionally, many detection frameworks are trained on specific generation methods and exhibit limited robustness to compression, blur, or unseen model distributions.

\subsection{Diffusion Model Analysis and Forensics}  
Diffusion models, grounded in reverse Markov chains and stochastic differential equations, have become the dominant paradigm for high-fidelity image synthesis~\cite{yang2023diffusionSurvey, chen2024opportunitiesDiffusion}. Recent studies explore their analytical properties: some demonstrate that pre-trained diffusion models can act as implicit detectors of synthetic imagery via strategic sampling~\cite{wang2024implicit_diffusion_detector}, while others investigate reconstruction error behavior in the latent space of diffusion decoders~\cite{jiang2025reconstruction_geometric}. Despite these promising directions, systematic forensic frameworks that exploit diffusion-model behavior—such as reconstruction dynamics or manifold alignment—remain limited.

\subsection{Manifold-Based Forensics and Feature Design}  
Traditional image forensics relies on pixel- or frequency-level cues, including sensor noise and compression artifacts~\cite{pentland1994statistics}. More recently, manifold-based reasoning has gained attention, suggesting that generative models learn implicit data manifolds where synthetic images lie on or near the manifold, while real images may deviate~\cite{song2024manifpt,jiang2025reconstruction_geometric}. Some approaches leverage reconstruction or projection errors in latent spaces; however, many require costly inversion procedures or assume knowledge of the generation model. As a result, generalization across unseen architectures and content types remains challenging.

\subsection{Research Gap and Motivation}  
Despite substantial progress in GAN detection and diffusion model analysis, key limitations persist. Existing detectors often lack cross-model generalization and robustness to real-world distortions. Many approaches rely on static image-level artifacts rather than analyzing dynamic reconstruction behavior. Furthermore, few methods treat diffusion models themselves as forensic probes for diverse content types beyond faces. These gaps motivate the development of dynamically informed features based on diffusion reconstruction trajectories that generalize across domains and generative architectures.

\section{Methodology}

\label{sec:method}

Our detection approach analyzes reconstruction behavior under varying noise levels in diffusion models. For convenience, we refer to this behavior as \textit{diffusion snap-back}.

\subsection{Dataset}

\begin{figure}[!t]
\centering
\newcommand{\imgheight}{1.8cm}

\includegraphics[height=\imgheight]{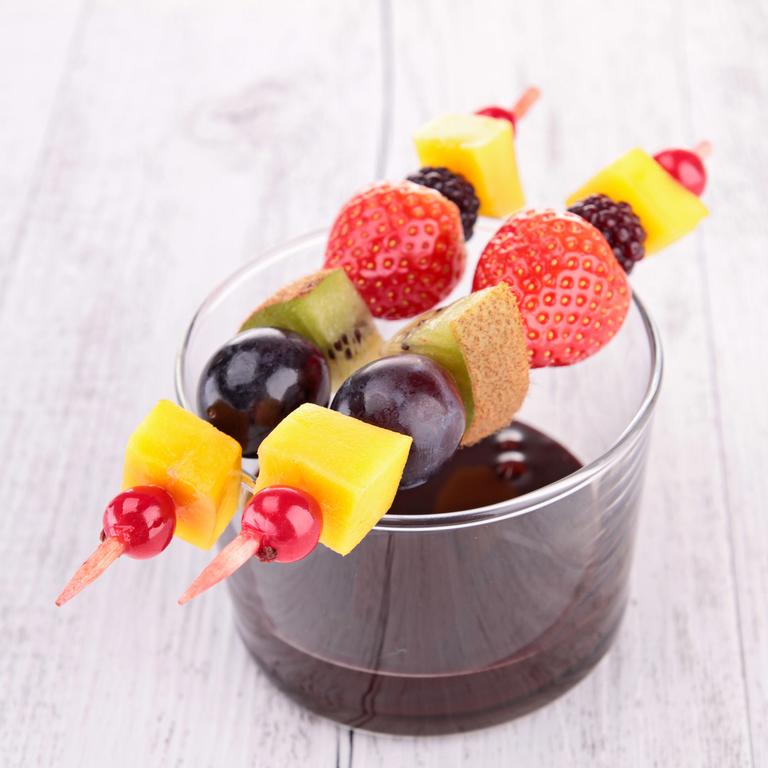}\hspace{0.25em}
\includegraphics[height=\imgheight]{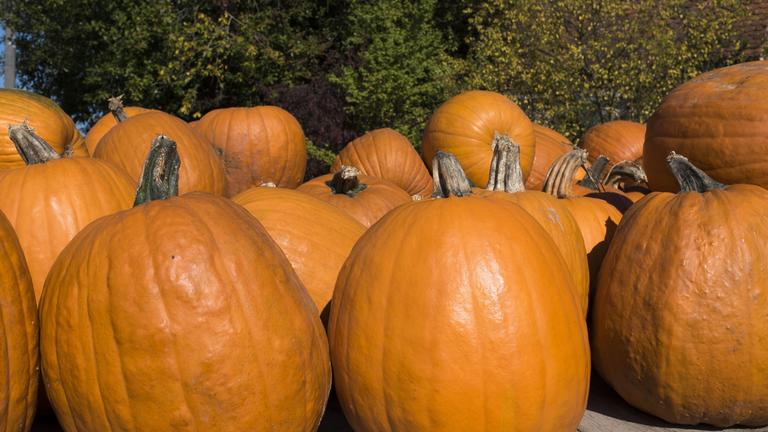}\hspace{0.25em}
\includegraphics[height=\imgheight]{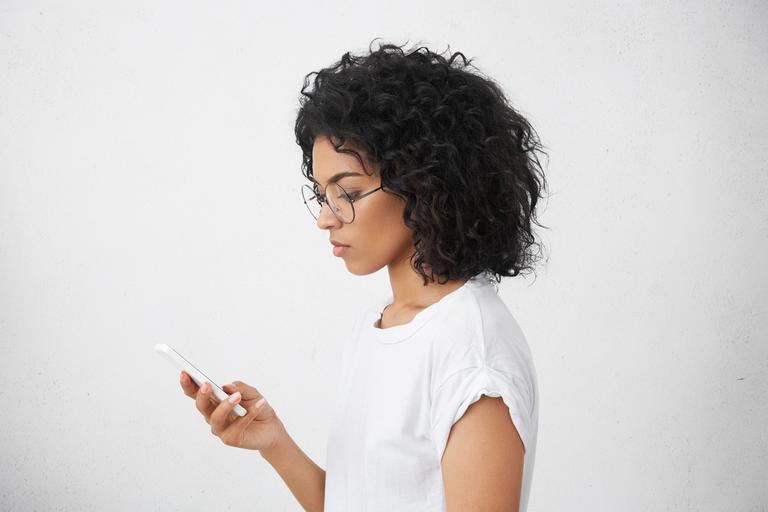}

\vspace{0.25em}

\includegraphics[height=\imgheight]{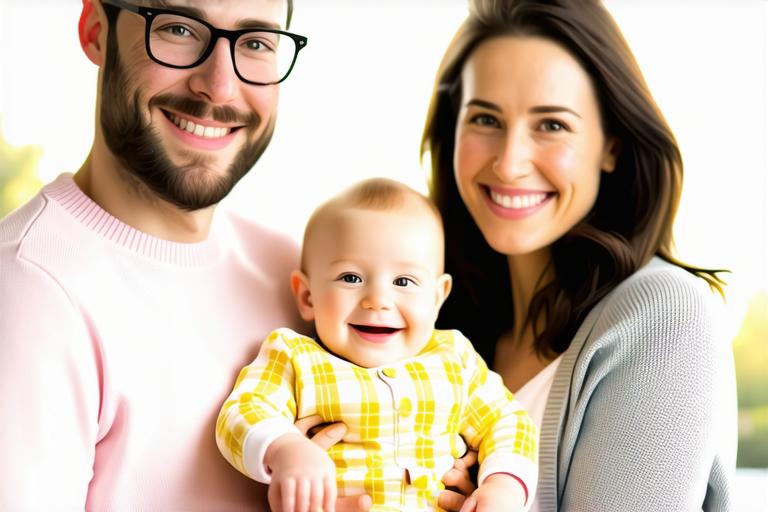}\hspace{0.25em}
\includegraphics[height=\imgheight]{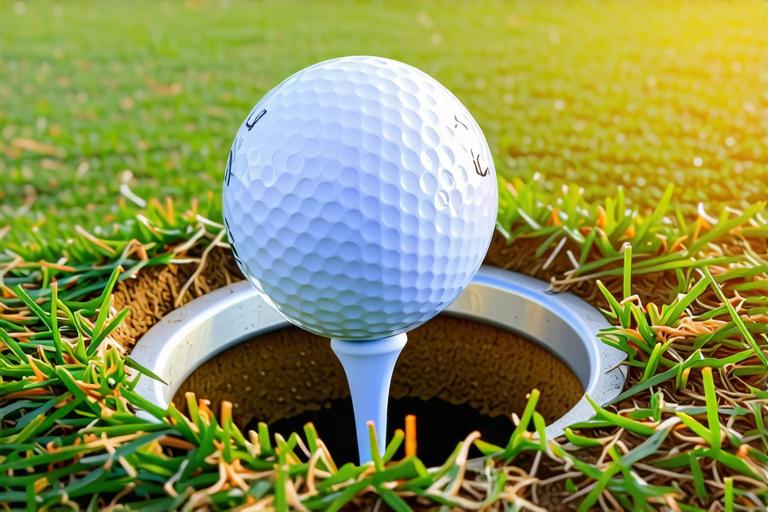}\hspace{0.25em}
\includegraphics[height=\imgheight]{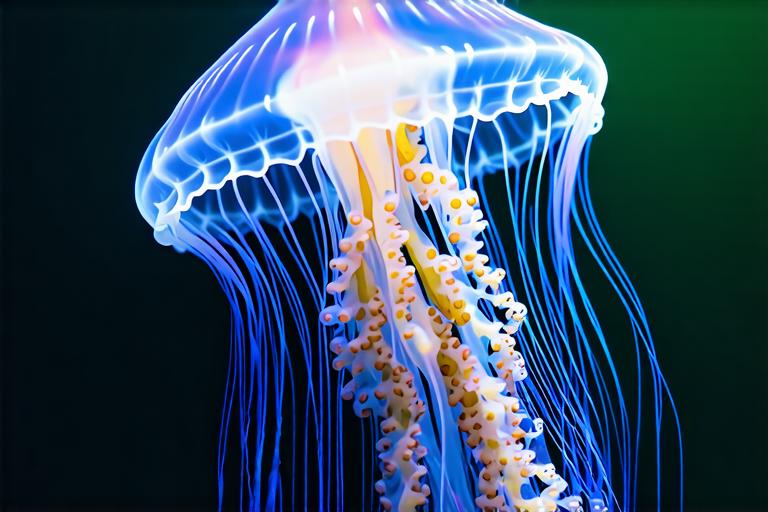}

\caption{Top row: \textbf{real (human-captured)} images. Bottom row: \textbf{AI-generated (synthetic)} images from the dataset.}
\label{fig:real_vs_fake_samples}
\end{figure}

\begin{table}[!t]
\caption{This is the sample dataset for training that contains label 0 (human-captured image) and 1 (AI-generated) image.}
\label{tab:sample_dataset}
\centering
\footnotesize
\begin{tabular}{l c}
\toprule
\textbf{File Name} & \textbf{Label} \\
\midrule
train\_data/a6dcb93f596a43249135678dfcfc17ea.jpg & 1 \\
train\_data/041be3153810433ab146bc97d5af505c.jpg & 0 \\
train\_data/615df26ce9494e5db2f70e57ce7a3a4f.jpg & 1 \\
train\_data/8542fe161d9147be8e835e50c0de39cd.jpg & 0 \\
train\_data/5d81fa12bc3b4cea8c94a6700a477cf2.jpg & 1 \\
\bottomrule
\end{tabular}
\end{table}

We used the \textit{AI vs. Human-Generated Images Dataset} by Alessandra Sala, publicly available on Kaggle~\cite{kaggle_alessandra_2024}, as the primary dataset for our experiments. The dataset contains equal proportions of authentic (human-captured) and AI-generated images across multiple categories, including portraits, objects, and scenes. Each image is labeled as \textbf{0} (human-generated) or \textbf{1} (AI-generated), as shown in Table~\ref{tab:sample_dataset}. Representative samples are presented in Figure~\ref{fig:real_vs_fake_samples}, where the top row shows real images and the bottom row shows AI-generated counterparts. Although the dataset is primarily face-centric, it includes diverse content such as portraits, landscapes, and objects, supporting evaluation across multiple domains. Authentic samples include photographs and original artwork, while synthetic samples are generated using Stable Diffusion v1.5.

\subsection{Diffusion Reconstruction and Feature Extraction}
Each input image $x$ is reconstructed through the Stable Diffusion img2img pipeline using a DDIM scheduler with 50 inference steps and guidance scale $w=1.0$. Four reconstruction strengths $\mathcal{S} = \{0.15, 0.30, 0.60, 0.90\}$ were applied to simulate increasing noise perturbation. For each strength, we computed three perceptual similarity metrics between the original image and its reconstruction: LPIPS (Learned Perceptual Image Patch Similarity, AlexNet backbone), SSIM (Structural Similarity Index), and PSNR (Peak Signal-to-Noise Ratio). This yielded 12 point-wise features per image. To capture global trajectory behaviour, three curve-level descriptors were derived from the metric profiles across $\mathcal{S}$: (1) AUC-LPIPS, the area under the LPIPS curve obtained via trapezoidal integration; (2) $\Delta_{LP}$, the difference between LPIPS values at $s_{\text{min}}=0.15$ and $s_{\text{mid}}=0.60$; and (3) the knee-step, defined as the first strength $s_*$ where SSIM dropped below $\tau = 0.80$. Together, these 15 features encode both local and global reconstruction dynamics, serving as interpretable indicators of manifold membership.

\subsection{Classification Pipeline}

\begin{figure}[!b]
  \centering
  \begin{tikzpicture}[
    node distance=0.4cm and 0pt,   
    every node/.style={
      rectangle, draw, rounded corners,
      text centered, font=\scriptsize,
      inner sep=2pt, minimum height=0.7cm, align=center
    },
    arrow/.style={thick, -{Stealth}}
  ]
    \node (A) {Input Image \\ (Real or AI)};
    \node (B) [below=of A] {Stage 1: Preprocessing \\ \textit{Resize 512×512 · Crop · Tensorize}};
    \node (C) [below=of B] {Stage 2: Diffusion Img2Img \\ \textit{Stable Diffusion v1.5 + DDIM} \\ Strengths \{0.15, 0.30, 0.60, 0.90\}};
    \node (D) [below=of C] {Stage 2a: Compute Metrics \\ \textit{LPIPS · SSIM · PSNR (per strength)}};
    \node (E) [below=of D] {Stage 3: Curve-Level Features \\ \textit{AUC-LPIPS · $\Delta$LP · Knee-Step}};
    \node (F) [below=of E] {Stage 4: Feature Matrix \\ \textit{15 Features → Clean + Scale}};
    \node (G) [below=of F] {Stage 5: Logistic Regression \\ \textit{Train + CV + Holdout}};
    \node (H) [below=of G] {Output: Binary Classification \\ \textit{0 = Real · 1 = Synthetic}};
    \node (I) [below=of H] {Robustness Tests \\ \textit{JPEG · WebP · Blur · Noise · Screenshot}};

    \draw [arrow] (A) -- (B);
    \draw [arrow] (B) -- (C);
    \draw [arrow] (C) -- (D);
    \draw [arrow] (D) -- (E);
    \draw [arrow] (E) -- (F);
    \draw [arrow] (F) -- (G);
    \draw [arrow] (G) -- (H);
    \draw [arrow] (H) -- (I);
  \end{tikzpicture}
  \caption{High-level flow of the synthetic vs real image classification pipeline.}
  \label{fig:flow_pipeline}
\end{figure}

The overall detection pipeline is illustrated in Figure~\ref{fig:flow_pipeline}. The extracted features were used to train a lightweight logistic regression classifier with $\ell_2$ regularization. Prior to training, missing values were imputed using the median and features were standardized. Stratified five-fold cross-validation was conducted to assess generalization, yielding mean AUROC and AUPRC values. The optimal decision threshold $\theta^*$ was determined via Youden’s J-statistic, balancing sensitivity and specificity. To benchmark performance, a pixel-level baseline using 32×32 flattened image vectors achieved only 0.525 AUROC, underscoring the effectiveness of the proposed manifold features. On the full feature set, the snap-back model reached 0.993 cross-validation AUROC and 0.990 on a 35\% holdout test split.

\subsection{Robustness Evaluation}

To assess real-world robustness, we evaluated the model on six augmentation conditions applied to a balanced subset of 24 images (12 real, 12 AI-generated): JPEG-60, WebP-60, Gaussian blur (radius 1.2), additive noise ($\sigma=6.0$), screenshot resampling, and raw images. Snap-back features were recomputed for each variant and AUROC was measured. As shown in Table~\ref{tab:robustness}, performance remained stable under compression (83–87\% AUROC), with moderate degradation under blur and screenshot distortions.

These results suggest that diffusion snap-back features retain discriminative power under common online perturbations.

\subsection{Implementation Details and Hyperparameters}
\label{subsec:hyperparams}

All experiments were conducted in Python using PyTorch, the diffusers library, lpips, and scikit-learn. Images were first corrected for orientation, converted to RGB format, center-cropped, and resized to 512×512 to ensure consistent preprocessing.

Reconstructions were generated using the Stable Diffusion v1.5 img2img pipeline with a DDIM scheduler and 50 denoising steps. Four reconstruction strengths (0.15, 0.30, 0.60, and 0.90) were applied to simulate increasing levels of controlled perturbation. Mixed-precision inference was used on GPU to improve computational efficiency.

For each reconstruction strength, three perceptual similarity measures—LPIPS (AlexNet backbone), SSIM, and PSNR—were computed between the original and reconstructed images, resulting in 12 direct comparison features. To summarize reconstruction behavior across strengths, we derived three additional descriptors: the area under the LPIPS curve, the change in LPIPS between low and moderate strengths, and the first reconstruction level at which structural similarity dropped below a threshold of 0.80. Together, these formed a compact 15-dimensional feature representation.

\section{Results and Discussion}

\begin{figure*}[!t]
    \centering
    \includegraphics[width=0.95\textwidth]{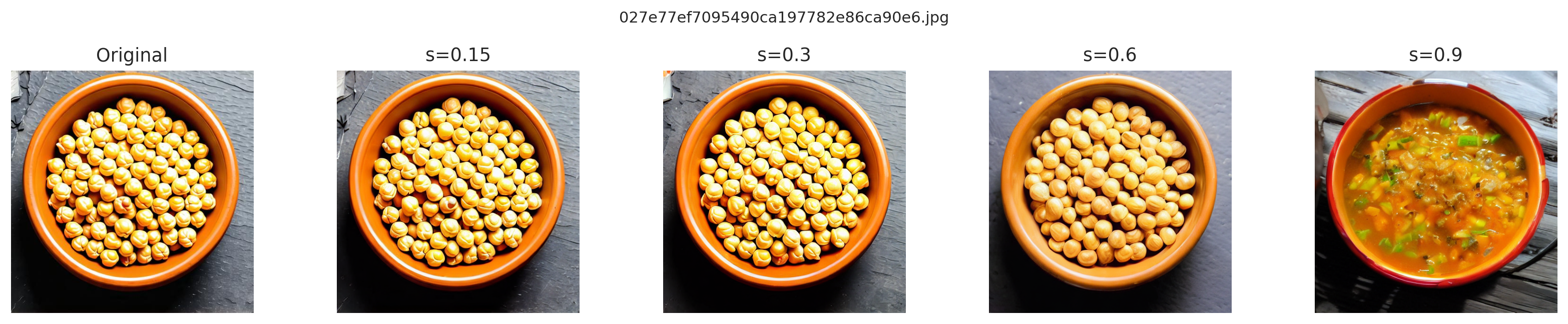}
    \caption{\textbf{AI-generated example (chickpeas bowl).}
    Progressive diffusion reconstructions at strengths $s=\{0.15,0.30,0.60,0.90\}$.
    The synthetic image remains visually consistent and semantically coherent even at $s=0.9$,
    showing smooth degradation characteristic of \textit{on-manifold} behavior.}
    \label{fig:grid_ai}
\end{figure*}

\begin{figure*}[!t]
    \centering
    \includegraphics[width=0.95\textwidth]{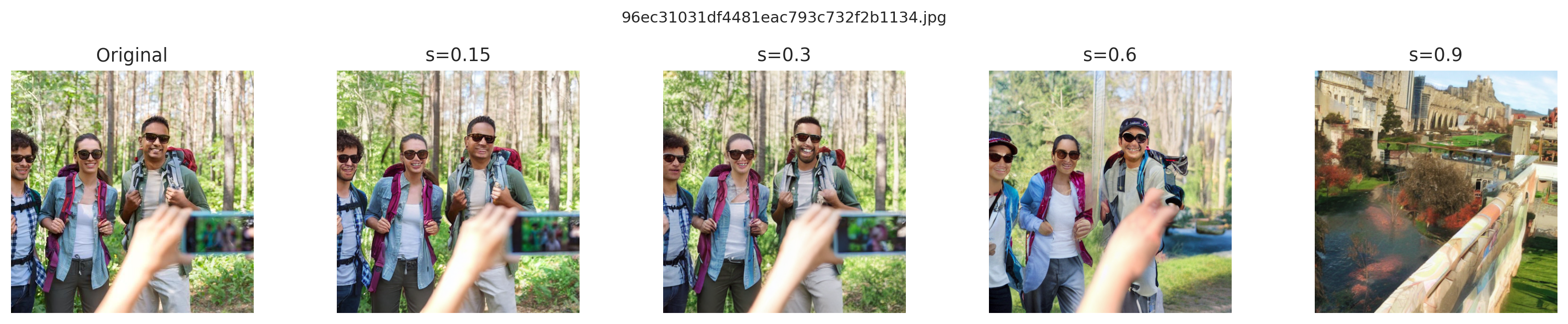}
    \caption{\textbf{Human-captured example (hikers group photo).}
    Authentic photographs exhibit strong off-manifold divergence at higher noise strengths—
    fine details and spatial coherence collapse rapidly beyond $s=0.6$,
    illustrating the \textit{knee-step} degradation pattern typical of real images.}
    \label{fig:grid_human}
\end{figure*}

\label{sec:results}

\subsection{Primary Results}

\begin{table}[!t]
\centering
\caption{Primary Detection Performance (5-fold CV, Full Dataset)}
\label{tab:primary}
\begin{tabular}{l|c|c|c}
\toprule
\textbf{Condition} & \textbf{AUROC} & \textbf{95\% CI} & \textbf{AUPRC} \\
\midrule
Full Dataset (CV) & $0.993$ & $[0.992, 0.994]$ & $0.991$ \\
Clean Subset (CV) & $0.824$ & $[0.767, 0.868]$ & $0.823$ \\
Test Holdout (35\%) & $0.990$ & -- & $0.988$ \\
\bottomrule
\end{tabular}
\end{table}

Table~\ref{tab:primary} summarizes the primary detection performance. Our diffusion snap-back feature pipeline achieves an \textbf{AUROC of 0.993} on 5-fold cross-validation, and maintains \textbf{0.990} on a 35\% holdout test set. The optimal decision threshold determined via Youden’s J-statistic is $\theta^* = 0.914$, yielding balanced sensitivity and specificity.  

\begin{figure*}[!t]
    \centering
    \includegraphics[width=0.95\textwidth]{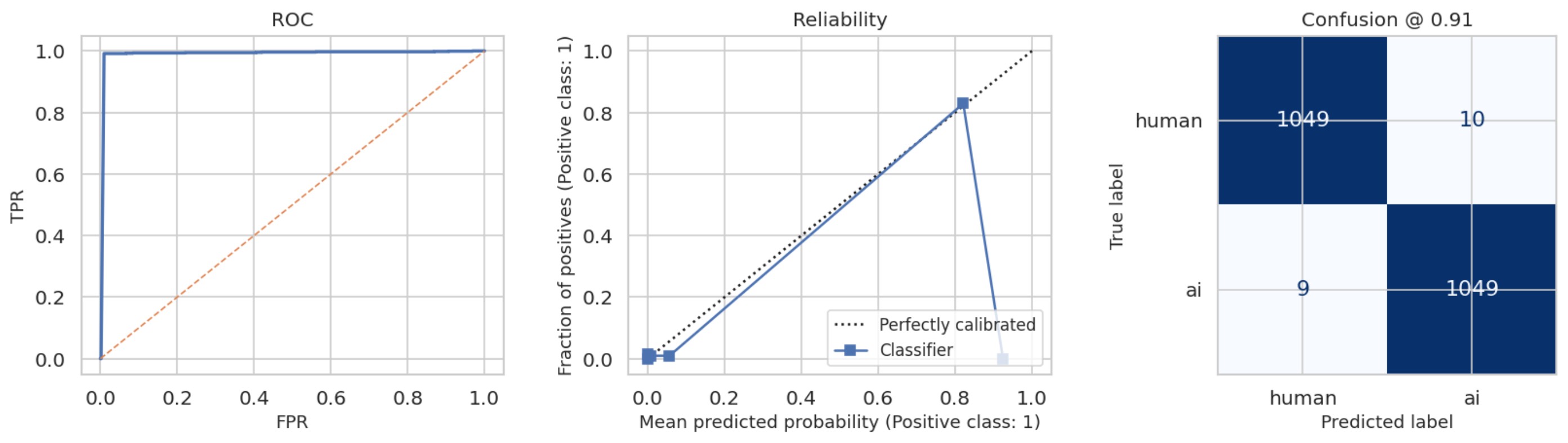}
    \caption{Evaluation metrics on the holdout set. (Left) ROC curve showing AUROC=$0.990$. (Middle) Reliability curve indicating close alignment with perfect calibration. (Right) Confusion matrix at $\theta^* = 0.914$ with minimal false positives/negatives.}
    \label{fig:roc_reliability_confusion}
\end{figure*}

The overall performance metrics are visualized in Fig.~\ref{fig:roc_reliability_confusion}, which shows the ROC curve, calibration reliability plot, and confusion matrix at $\theta^*$. The model demonstrates high discriminative power and near-perfect calibration, with minimal false positives or negatives.

\subsection{Ablation Study}

\begin{table}[!t]
\centering
\caption{Feature Ablation: Top 5 Combinations}
\label{tab:ablation}
\begin{tabular}{l|c}
\toprule
\textbf{Feature Set} & \textbf{CV AUROC} \\
\midrule
$\text{knee\_step} + \text{lpips\_0.6} + \text{auc\_lpips}$ & $0.987$ \\
$\text{ssim\_0.6} + \text{lpips\_0.15}$ & $0.978$ \\
$\text{lpips\_0.15} + \text{lpips\_0.6}$ & $0.976$ \\
$\text{auc\_lpips}$ (single) & $0.915$ \\
$\text{lpips\_0.6}$ (single) & $0.903$ \\
\bottomrule
\end{tabular}
\end{table}

Table~\ref{tab:ablation} presents the top-performing feature combinations. Ablation analysis identifies \textit{knee-step} (the threshold at which SSIM drops below 0.8) as the single most discriminative feature. Combined with LPIPS at higher diffusion strengths and curve-level AUC summaries, the model approaches full-feature accuracy with minimal redundancy.

\subsection{Metric Trajectories and Snap-Back Behavior}

\begin{figure*}[!t]
    \centering
    \includegraphics[width=0.32\textwidth]{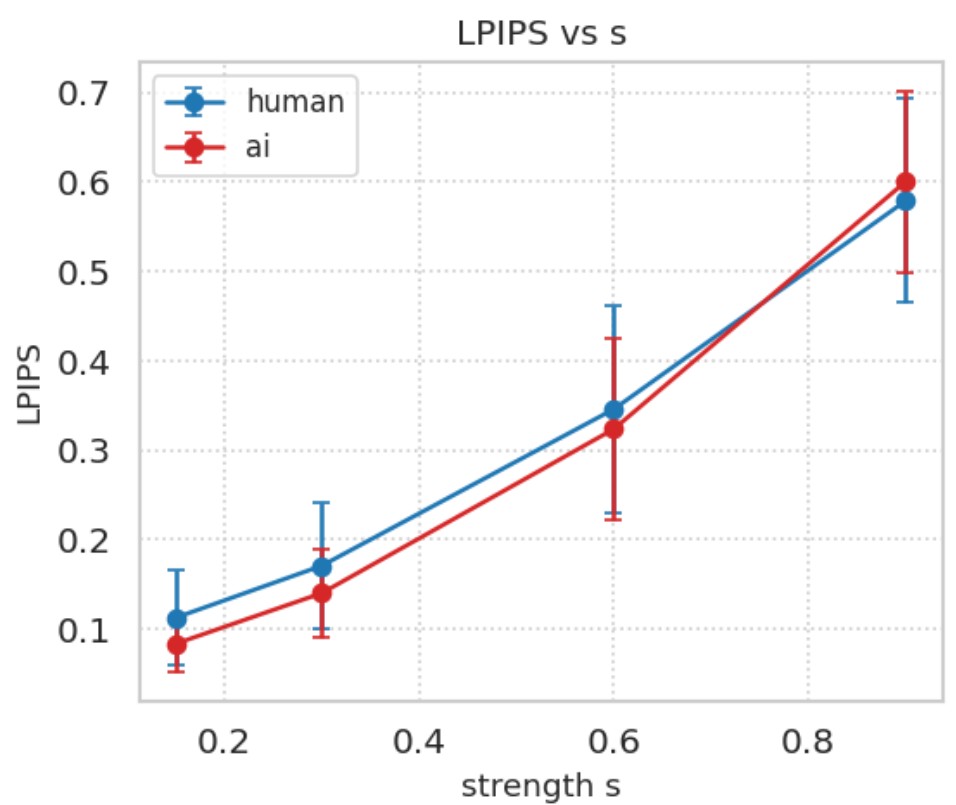}
    \hfill
    \includegraphics[width=0.32\textwidth]{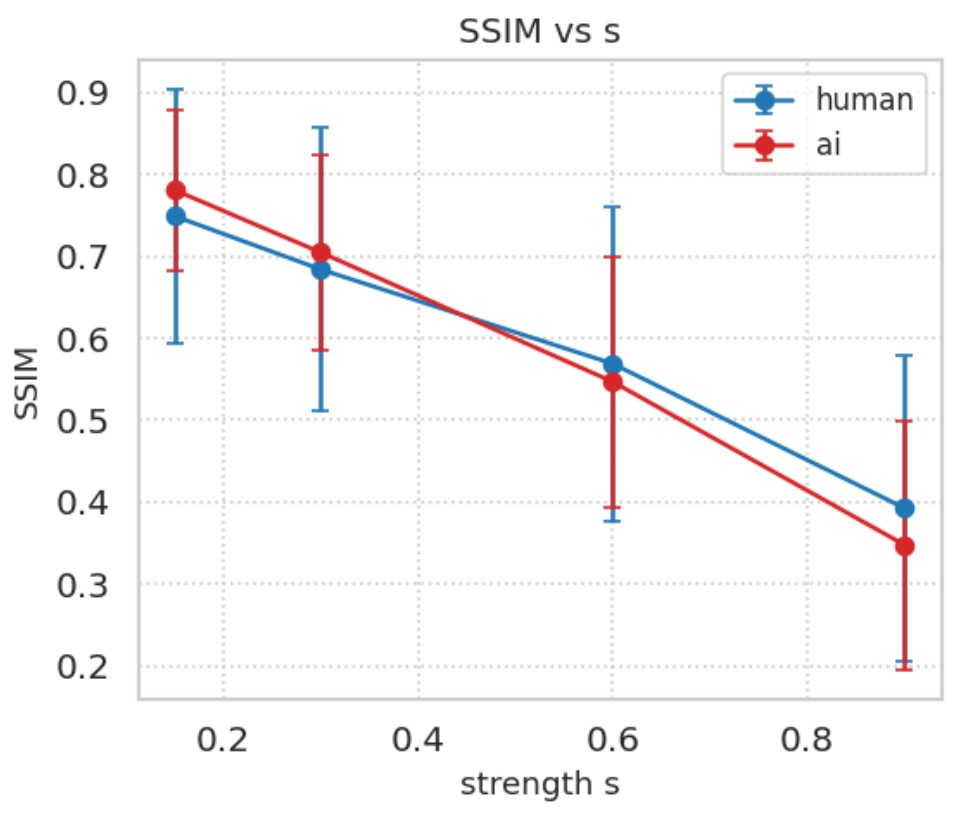}
    \hfill
    \includegraphics[width=0.32\textwidth]{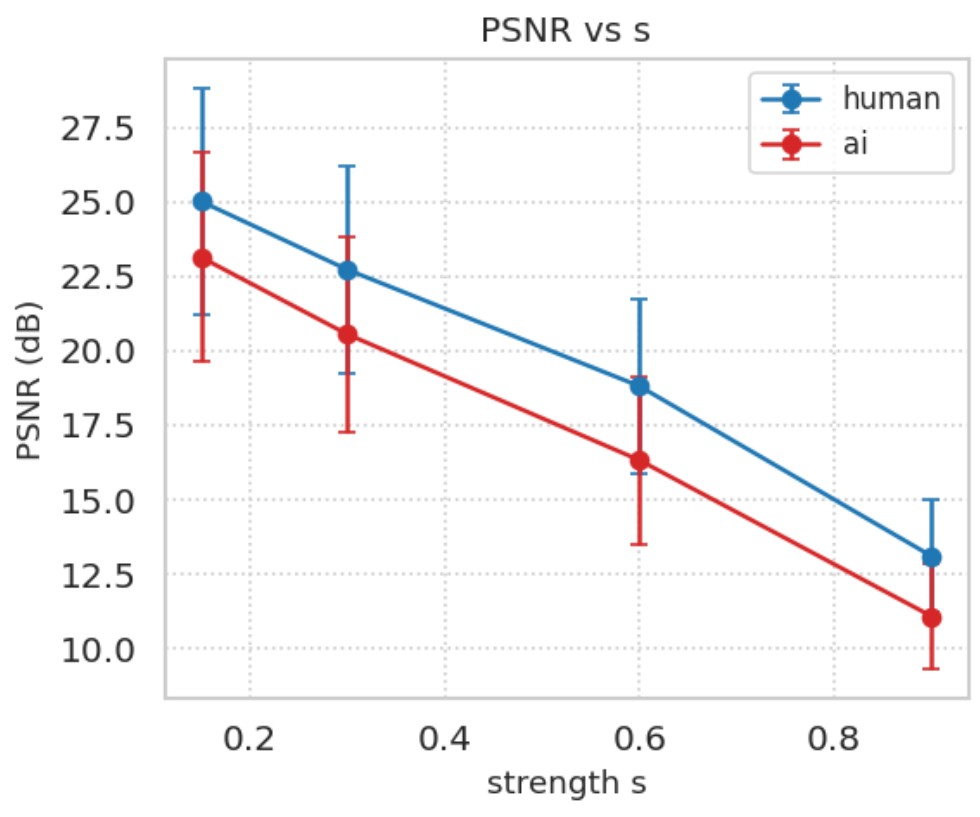}
    \caption{Metric trajectories across diffusion strengths: LPIPS (left), SSIM (middle), and PSNR (right). Human images exhibit abrupt degradation beyond $s>0.6$, while AI-generated images degrade smoothly, reflecting manifold proximity.}
    \label{fig:metric_trends}
\end{figure*}

The diffusion reconstruction trajectories of LPIPS, SSIM, and PSNR are illustrated in Fig.~\ref{fig:metric_trends}, highlighting the characteristic degradation differences between human and AI-generated images. For human (blue) vs. AI (red) images, human samples show sharper degradation patterns—particularly a steeper LPIPS increase and PSNR decay—consistent with their off-manifold reconstructions.

\begin{figure}[!t]
    \centering
    \includegraphics[width=0.90\columnwidth]{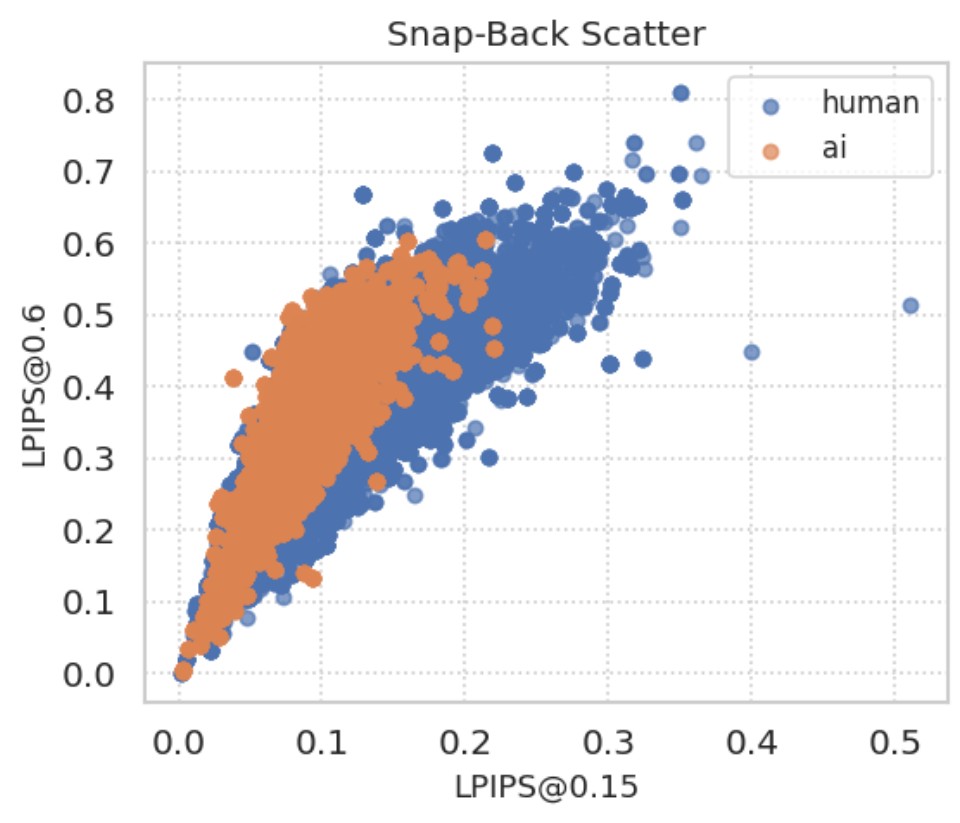}
    \caption{Scatter of LPIPS@$s{=}0.15$ vs. LPIPS@$s{=}0.6$. AI (orange) and human (blue) clusters exhibit distinct separation, supporting low-dimensional feature discriminability.}
    \label{fig:snapback_scatter}
\end{figure}

The joint scatter in Figure~\ref{fig:snapback_scatter} illustrates separability between classes based on LPIPS at weak ($s=0.15$) vs. moderate ($s=0.6$) noise strengths, showing almost linear separability in 2D feature space.

\subsection{Feature Correlations}

\begin{figure}[!t]
\centering
\includegraphics[width=0.9\columnwidth]{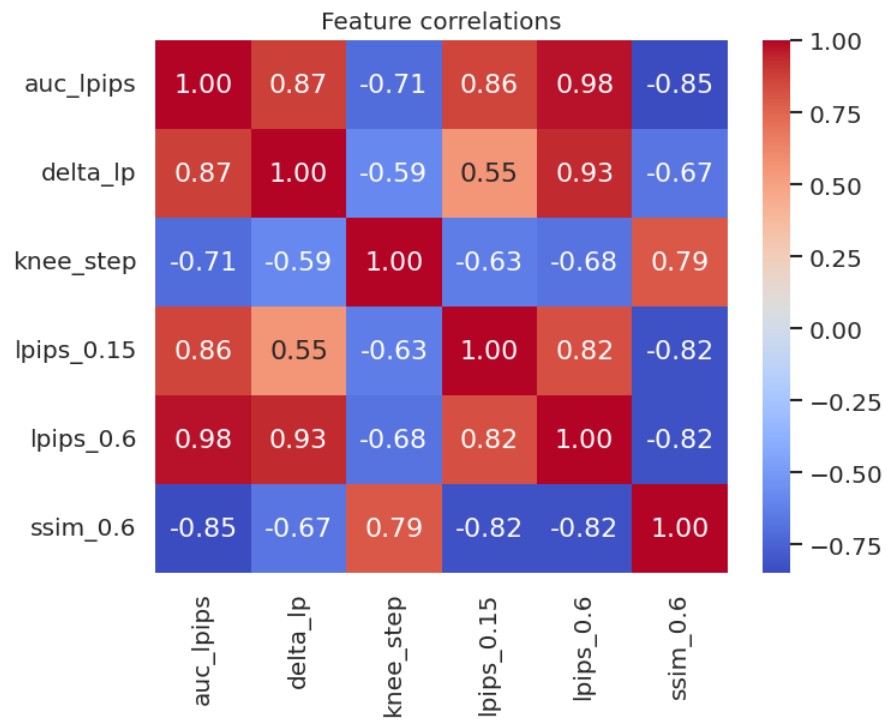}
\caption{Feature correlation heatmap showing complementarity of global (AUC-LPIPS, knee-step) and local (LPIPS@0.15, LPIPS@0.6, SSIM@0.6) features.}
\label{fig:corr_heatmap}
\end{figure}

The inter-feature dependencies are shown in Fig.~\ref{fig:corr_heatmap}, where the heatmap reveals the complementarity of global (AUC-LPIPS, knee-step) and local (LPIPS@0.15, LPIPS@0.6, SSIM@0.6) metrics.

\subsection{Qualitative Visualization}

Figures~\ref{fig:grid_ai} and~\ref{fig:grid_human} present qualitative reconstructions illustrating the snap-back process for AI-generated and human-captured samples. AI-generated images remain semantically coherent and visually stable even under high diffusion noise, whereas authentic images diverge sharply and lose structural consistency beyond $s=0.6$.

\subsection{Robustness Evaluation}

\begin{table}[!t]
\centering
\caption{Per-Augmentation Robustness (24-image pilot)}
\label{tab:robustness}
\begin{tabular}{l|c}
\toprule
\textbf{Augmentation} & \textbf{AUROC} \\
\midrule
Raw & $0.833$ \\
JPEG-60 & $0.833$ \\
WebP-60 & $0.867$ \\
Blur & $0.700$ \\
Noise & $0.800$ \\
Screenshot & $0.767$ \\
\bottomrule
\end{tabular}
\end{table}

Lossy compression (JPEG/WebP) minimally impacts classification accuracy (83–87\% AUROC), while geometric or spatial distortions (blur, screenshot) reduce detection to 70–77\%. Interestingly, WebP compression slightly improves separability, potentially accentuating generative artifacts.

\subsection{Why Snap-Back Works}

Diffusion models learn a denoising function $\nabla \log p_\theta(x)$ that approximates the data distribution through score matching. Images generated by the model are strongly aligned with this learned prior, whereas real-world photographs may occupy regions that are less precisely represented. 

When noise is injected and reconstruction begins, this difference becomes apparent. Real images often diverge more sharply under high noise, resulting in larger LPIPS increases and faster SSIM degradation. In contrast, AI-generated images tend to recover more smoothly, maintaining structural consistency across reconstruction strengths. 

This contrast in reconstruction dynamics reflects varying degrees of alignment with the learned denoising prior. As such, diffusion snap-back provides an interpretable signal for distinguishing synthetic images from authentic ones.

\subsection{Deployment Considerations}
\label{sec:deployment}

The snap-back framework is lightweight and modular, requiring four img2img reconstructions followed by metric computation and a linear classifier, enabling transparent and efficient inference.

It can be deployed as an ``upload-and-check'' module within admission portals, recruitment systems, journalism workflows, or social media platforms to flag potentially synthetic submissions. Inference can be accelerated through batching and GPU parallelization, and the framework can be adapted to alternative diffusion backbones with minimal modification.

Overall, the separation between reconstruction dynamics and linear classification supports practical and interpretable real-world deployment.

\section{Conclusion}
\label{sec:conclude}

This work introduced a diffusion-based forensic framework for distinguishing authentic photographs from AI-generated images by analyzing their reconstruction behavior under controlled noise strengths. Rather than relying on static pixel-level artifacts, we modeled multi-strength img2img reconstruction and extracted interpretable snap-back features aligned with the diffusion model’s learned denoising prior. The proposed approach demonstrates that reconstruction dynamics can serve as a complementary and principled signal for synthetic media detection.

Several limitations remain. Experiments were conducted using a single diffusion backbone, and broader cross-model validation is necessary to assess generalization. Hyperparameter exploration was limited, and more systematic tuning of reconstruction strengths and thresholds may further refine performance. Additionally, larger and more diverse datasets are required to better understand robustness across domains.

Future work will focus on cross-diffusion evaluation, expanded hyperparameter optimization, and scaling reconstruction analysis with improved computational efficiency. Extending snap-back analysis to video also represents a promising direction for advancing reliable and scalable synthetic media forensics.

\bibliography{ref}
\bibliographystyle{IEEEtran}

\end{document}